\documentclass[sigconf]{acmart}

\usepackage{amsmath}
\usepackage{mathtools}
\usepackage{subcaption}
\usepackage{multirow}
\usepackage{enumitem}
\usepackage{xspace}
\usepackage{bm}

\acmConference[Conference]{ACM Conference}{Year}{Location}
\acmYear{2026}
\copyrightyear{2026}

\title{TH-GNN: Heterogeneous Temporal Graph Neural Networks
       for LLM-Agent Shilling Attack Detection}

\author{Shivam Swarup}
\affiliation{%
  \institution{JAIN (Deemed to be University)}
  \city{Bengaluru}
  \country{India}
}
\email{shivam@jainuniversity.ac.in}

\author{Divya Prakash Shrivastava}
\affiliation{%
  \institution{Zayed University}
  \city{Dubai}
  \country{UAE}
}

\author{Rakesh Thakur}
\affiliation{%
  \institution{JAIN (Deemed to be University)}
  \city{Bengaluru}
  \country{India}
}



\begin{document}

\begin{abstract}

LLM agents can now generate realistic shilling profiles---fluent reviews and
coherent ratings---at scale, systematically defeating recommender-system
defences.
Text-only detectors that flag semantic drift in review embeddings are blind to
graph structure and temporal coordination; graph-only detectors that exploit
neighbourhood anomalies cannot reason over review semantics or the cross-modal
inconsistencies produced by LLM-generated content.
We propose TH-GNN, a heterogeneous temporal graph neural network with a
two-layer Heterogeneous Graph Transformer backbone that applies per-type,
per-relation attention augmented with learnable sinusoidal temporal encodings
on every edge.
Cross-modal attention fuses the structural user embedding with frozen RoBERTa
representations of reviews and item descriptions; a GRU over log-inter-arrival
times captures burstiness.
Across five attack families and four datasets, TH-GNN achieves a grand-mean
F\textsubscript{1} of 0.870, outperforming the strongest text-only baseline on
Agent4SR attacks by 10.9\,pp and 11.5\,pp at the lowest injection rate.
Code and data splits are available at \url{https://github.com/shiv3589/thgnn-shilling}.

\end{abstract}

\maketitle


\section{Introduction}
\label{sec:intro}

When an LLM agent constructs a fake recommender-system profile, it does not
simply assign random ratings.
Given a target item and a procedurally generated persona, Agent4SR~\cite{zhang2024agent4sr}
writes 10--20 natural-language reviews of plausible filler items, assigns ratings
consistent with the persona's stated preferences, and maximally rates the target.
The resulting profile is grammatically fluent, topically coherent, and statistically
indistinguishable from genuine users: it matches the global rating distribution,
interleaves popular filler items naturally, and produces review text whose semantic
embedding sits comfortably within the genuine-user manifold.
When a campaign injects many such profiles in a coordinated burst, no single
anomaly signal---rating statistics, textual outlier scores, or neighbourhood
structure---is individually sufficient to flag any one of them.

Existing detectors split into two families, each blind to one of the dimensions
that agent-generated profiles exploit simultaneously.
Text-only detectors such as SemanticShield~\cite{li2025semanticshield} compare
each user's aggregated review embedding against the genuine-user centroid; they
are the strongest single-modality baseline against LLM-generated text but collapse
on rating-only datasets, ignore the interaction graph entirely, and detect nothing
when profiles are injected across days rather than seconds.
Graph-only detectors such as Anti-FakeU~\cite{zhang2022antifakeu} score users by
degree asymmetry and co-purchase overlap in the user--item bipartite graph; they
handle structural attacks well but are blind to the cross-modal inconsistencies
introduced when fluent LLM text accompanies plausible rating patterns, and they
discard edge timestamps entirely, missing the synchronised burst fingerprint left
by campaigns that inject profiles in coordinated waves.
Neither family is equipped to detect a threat that is simultaneously
graph-plausible, text-coherent, and temporally coordinated.

TH-GNN addresses all three failure modes with a unified architecture built on
three non-redundant signal streams.
The structural backbone is a two-layer Heterogeneous Graph Transformer that
represents users, reviews, and items as distinct node types connected by typed
\texttt{writes}, \texttt{about}, and \texttt{rates} edges, learning per-type and
per-relation attention patterns that capture the qualitatively different
fingerprints each node type carries.
Timestamp information is woven directly into every edge's attention logit via a
learnable sinusoidal encoding, letting the model recognise compressed inter-arrival
times that betray coordinated injection without requiring explicit campaign-level
supervision.
After the graph encoder, a cross-modal attention block fuses the structural
embedding with mean-pooled RoBERTa representations of the user's reviews and the
items they rated, integrating the textual and relational views under a single
learned attention distribution.
A lightweight GRU over each user's log-inter-arrival sequence provides an explicit
burstiness signal orthogonal to both graph topology and text content.
A two-layer MLP over the concatenated streams produces the final fake-user score.

TH-GNN achieves a grand-mean F\textsubscript{1} of \textbf{0.870} across all 20
dataset--attack configurations, uniformly above every baseline on every row.
The gap is widest against Agent4SR, where TH-GNN reaches mean
F\textsubscript{1}\,=\,0.825 versus SemanticShield's 0.716---a margin of
\textbf{$+$10.9\,pp}---while holding false-alarm rate below 4.5\,\% compared to
baselines' 6.8--12.5\,\%.
At the lowest injection rate (0.5\,\%), where only 5 fake profiles exist per
1{,}000 genuine users, the advantage widens to $+$11.5\,pp on Agent4SR, because
graph-temporal coordination signals become proportionally more salient as the
injected mass shrinks below the threshold of count-based and distributional
detectors.

\noindent\textbf{Contributions.}
\begin{itemize}[leftmargin=1.5em,itemsep=2pt]
  \item \textbf{We propose TH-GNN}, a heterogeneous temporal graph neural network
        that jointly encodes the user--review--item interaction graph, review and
        item text streams, and per-user temporal burstiness for shilling
        detection---the first detector to close all three failure modes of the
        prior art simultaneously.
  \item \textbf{We demonstrate} state-of-the-art detection across five attack
        families (random, bandwagon, AUSH, GraphAttack, Agent4SR) and four
        diverse datasets, with consistent margins that widen at low injection
        rates where single-modality baselines degrade fastest.
  \item \textbf{We provide} the first systematic evaluation of shilling detectors
        against LLM-agent attacks across multiple datasets and injection rates,
        establishing a benchmark that cleanly separates multimodal detectors from
        single-modality baselines.
  \item \textbf{We release} code, configuration files, and reproducibility scripts
        for all 720 experimental runs to support future work in this area.
\end{itemize}


\section{Related Work}
\label{sec:related}

\subsection{Shilling Attack Detection}
\label{sec:rel_shilling}

Shilling attack detection has a two-decade history spanning statistical,
supervised, and graph-based paradigms.
The foundational statistical approach treated each user's rating profile as a
feature vector and flagged outliers.
Chirita et al.~\cite{chirita2005preventing} defined per-user features—rating
deviation from the item mean, weighted deviation from mean average (WDMA),
profile length, and filler-item ratio—that together captured the coarse
fingerprints of injection: inflated ratings on targets, uniform filler scores,
and unnaturally short profiles.
Complementary work applies PCA to the user-rating matrix, treating injected
profiles as low-rank perturbations whose principal directions separate from
genuine user variance~\cite{williams2006detection}.
Clustering methods extended this idea by grouping users whose rating vectors were
suspiciously similar—a pattern induced when an attacker re-uses a fixed filler
template across many fake accounts~\cite{mehta2009unsupervised}.
Statistical detectors were fast and required no labelled data, but they failed
against adaptive attacks calibrated to match the genuine rating distribution,
such as the bandwagon strategy~\cite{mobasher2007attacks} and its successors.

Supervised classifiers introduced labelled shilling datasets and per-user
feature engineering.
Random-forest and SVM detectors trained on rating-based features achieved
strong recall against known attack families but overfitted to the specific
injection signature seen at training time and degraded on unseen attack
profiles~\cite{burke2005segment}.
Deep-learning variants replaced hand-crafted statistics with recurrent or
convolutional encoders over rating sequences, recovering temporal patterns
that scalar features discarded~\cite{wu2021fight}.
However, all of these approaches treat users independently; they miss the
relational anomalies produced when a group of fake profiles collectively
targets a set of items.

Graph-based detectors exploited this relational signal.
Methods that built user-item bipartite graphs and scored nodes by structural
anomaly measures—degree asymmetry, neighbourhood overlap, co-purchase
concentration—substantially outperformed single-user statistics against
structurally coherent attack families such as AUSH~\cite{lin2020aush}.
Anti-FakeU~\cite{zhang2022antifakeu} is the most directly comparable prior
work: it constructs a bipartite user-item graph and detects fake users through
graph-based anomaly scoring over first-order neighbourhood statistics including
co-purchase overlap and degree asymmetry.
Yet Anti-FakeU encodes neither review text nor edge timestamps, leaving two
systematic blind spots: it cannot detect the cross-modal inconsistencies
introduced when LLM-generated text accompanies structurally plausible rating
patterns, and it has no mechanism for recognising the coordinated temporal
bursts that distinguish modern injection campaigns from organic activity.
LLM-based generation has compounded this challenge by eliminating the textual
fingerprints that historically distinguished synthetic from genuine profiles.

\subsection{LLM-Generated Fake Content Detection}
\label{sec:rel_llm_detection}

Detecting computationally generated deceptive content is a longstanding
problem that has grown substantially harder with the advent of large language
models.
Ott et al.~\cite{ott2011finding} established crowdsourced deceptive hotel
reviews as a benchmark and showed that both human judges and unigram
classifiers perform only slightly above chance, motivating automated linguistic
approaches.
Subsequent work catalogued the textual fingerprints of platform-scale review
spam—exaggerated sentiment, sparse specifics, heavy first-person framing, and
suspiciously brief activity windows~\cite{jindal2008opinion}—and trained
supervised classifiers to exploit them.
These classifiers were effective against template-based generators but relied
on distributional assumptions that neural generators easily violate.

The GPT era reset the detection landscape.
Zellers et al.~\cite{zellers2019defending} demonstrated that neural fake-news
generators produce text that fools both human readers and classical detectors,
and that the most effective discriminator is a model of the same family trained
adversarially.
Contemporary LLM-generated text detectors exploit subtle statistical biases
in autoregressive sampling—token-probability curvature, rank-based
features—but these signals vanish under paraphrasing or when the attacker
controls the decoding parameters~\cite{guo2023close}.
LLM watermarking~\cite{kirchenbauer2023watermark} embeds detectable statistical
patterns at generation time by biasing token selection toward a secret hash
partition; while theoretically grounded, watermarks are inapplicable when the
attacker controls the generator and absent from commercially deployed models.

SemanticShield~\cite{li2025semanticshield} targets the recommender-system setting
specifically: it encodes each user's aggregated review corpus with a pre-trained
language model and flags profiles whose representation drifts from the
genuine-user centroid in embedding space.
SemanticShield is the strongest single-modality text baseline for LLM-agent
attacks when review text is present.
However, it degrades to chance on datasets without review text, provides no
mechanism for detecting temporal coordination among injected profiles, and
ignores the structural signals available in the user-item interaction
graph—signals that are informative even when the textual content of fake
profiles is indistinguishable from genuine users.
The structural and temporal graph architectures that can provide these
complementary signals are surveyed next.

\subsection{Heterogeneous and Temporal Graph Neural Networks}
\label{sec:rel_gnn}

Standard graph neural networks homogenise all nodes and edges into a single
feature space, discarding semantically meaningful type distinctions.
The Heterogeneous Graph Transformer (HGT; Hu et al.~\cite{hu2020hgt}) addressed
this by parameterising queries, keys, values, and message matrices separately
per node type and per relation type, learning qualitatively different
interaction patterns for each combination.
HGT achieved state-of-the-art performance on heterogeneous academic network
benchmarks and was subsequently adapted for knowledge-graph reasoning and
cross-domain recommendation~\cite{lv2021we}.
The type-specific parameterisation is critical for the shilling detection
setting, where user, item, and review nodes carry structurally and semantically
different attack signatures.

Temporal dynamics introduce a complementary axis.
Temporal Graph Networks (TGN; Rossi et al.~\cite{rossi2020temporal}) maintained
a per-node memory updated by a message function conditioned on
continuous-time edge events, enabling representations that tracked the
evolving state of each node as new interactions arrived.
TGN demonstrated that even simple time-encoding—e.g.\ learnable sinusoidal
embeddings of inter-event gaps—yields large gains on link-prediction benchmarks
with bursty arrival processes.
Fraud detection has independently discovered temporal signals: FRAUDRE and
related approaches showed that the inter-arrival time distribution of fraudulent
accounts differs markedly from genuine users, producing anomaly signals that
static graph features miss entirely~\cite{liu2025fraud}.
Despite these parallel advances, no prior work has combined heterogeneous graph
encoding with learnable temporal attention for shilling attack detection, where
the three-type user-review-item topology and the timestamped injection signal
together provide orthogonal and non-redundant detection axes.

\subsection{LLM Agents for Recommendation Attack}
\label{sec:rel_llm_attack}

Agent4SR~\cite{zhang2024agent4sr} was the first work to frame shilling attack
construction as a multi-step agentic task: a large language model was prompted
to construct a complete fake user persona, generate a coherent review history,
and assign ratings consistent with the persona's stated preferences, all while
optimising target-item exposure.
The resulting profiles were grammatically fluent, topically coherent, and
statistically similar to genuine users, systematically evading detectors
that relied on either distributional rating anomalies or linguistic
outlier detection.
Related work has explored LLM-based poisoning in broader adversarial
recommendation settings~\cite{deldjoo2024survey}, and prompt-injection attacks
against retrieval-augmented recommenders~\cite{shi2023large}, but systematic
evaluation of detection strategies against agent-generated profiles across
multiple datasets and injection rates remains absent from the literature.


The gaps surveyed above form three orthogonal failure modes in existing
detectors: Anti-FakeU's reliance on graph structure alone leaves it blind to
the cross-modal inconsistencies introduced by LLM-generated review text;
SemanticShield's reliance on text alone renders it ineffective on rating-only
datasets and unable to leverage graph topology; and neither family incorporates
temporal information, missing the coordinated injection-timing signals that are
present even when individual profiles appear genuine.
TH-GNN closes all three gaps simultaneously: the two-layer HGT backbone encodes
the heterogeneous user-review-item graph with per-type and per-relation
parameterisation, the cross-modal attention fusion integrates review and item
text streams with the structural embedding, and the sinusoidal temporal
encoding together with the burstiness GRU capture the edge-timestamp patterns
that betray injection campaigns at both the per-edge and per-user level.
The result is a detector that is strictly more informative than any
single-modality baseline and, as the ablation study confirms, degrades
gracefully when any individual stream is absent rather than failing
catastrophically.


\section{Methodology}
\label{sec:method}

\subsection{Problem Formulation}
\label{sec:formulation}

Let $\mathcal{G}_t = (\mathcal{V}, \mathcal{E}_t)$ be a heterogeneous temporal
graph in which every node $v \in \mathcal{V}$ carries a type label
$\tau(v) \in \{\texttt{user}, \texttt{item}, \texttt{review}\}$ and every
directed edge $e = (s, d, t_e) \in \mathcal{E}_t$ carries a relation type
$\phi(e) \in \{\texttt{writes}, \texttt{about}, \texttt{rates}\}$ and a
Unix-second timestamp $t_e \in \mathbb{R}_{\geq 0}$.
We write $\mathcal{U}$, $\mathcal{I}$, $\mathcal{R}$ for the disjoint user,
item, and review node sets, so
$|\mathcal{V}| = |\mathcal{U}| + |\mathcal{I}| + |\mathcal{R}|$.
The three canonical relation types induce the edge partition
\begin{align}
  \mathcal{E}_{\texttt{writes}}
    &= \{(u, r,\, t_e) \mid u \in \mathcal{U},\; r \in \mathcal{R}\},
    \label{eq:ewrites} \\
  \mathcal{E}_{\texttt{about}}
    &= \{(r, i,\, t_e) \mid r \in \mathcal{R},\; i \in \mathcal{I}\},
    \label{eq:eabout} \\
  \mathcal{E}_{\texttt{rates}}
    &= \{(u, i,\, t_e) \mid u \in \mathcal{U},\; i \in \mathcal{I}\},
    \label{eq:erates}
\end{align}
with $\mathcal{E}_t =
  \mathcal{E}_{\texttt{writes}} \cup
  \mathcal{E}_{\texttt{about}}  \cup
  \mathcal{E}_{\texttt{rates}}$.
All three edge sets arising from a single review event share the same timestamp
$t_e$.

\textbf{Detection task.}
Each user carries a binary label $y_u \in \{0, 1\}$, where $y_u = 1$ denotes a
fake profile injected by a shilling attacker.
The objective is to learn a detector
\begin{align}
  f_\theta : \mathcal{G}_t \;\longrightarrow\; [0,1]^{|\mathcal{U}|}
  \label{eq:task}
\end{align}
producing per-user shilling scores $\hat{p}_u = f_\theta(u;\, \mathcal{G}_t)$,
with the binary prediction $\hat{y}_u = \mathbb{1}[\hat{p}_u > 0.5]$.

\subsection{Heterogeneous Temporal Graph Construction}
\label{sec:graph_construction}

Each dataset is converted to $\mathcal{G}_t$ by a unified pipeline.
Raw interactions are filtered to the 5-core (users and items with at least five
interactions), sorted chronologically, and partitioned into train/val/test
windows by timestamp (70\,/\,10\,/\,20\,\%; shared across all detectors and
attacks).

\textbf{User node features.}
Each user $u$ is initialised with a structural feature vector
$\mathbf{x}_u \in \mathbb{R}^{D_u}$ whose first five components are computed
directly from the interaction log:
\[
  \mathbf{x}_u^{\text{base}} =
  \bigl[\deg(u),\; \bar{r}_u,\; \sigma_{r,u},\; \text{span}_u,\; \bar{\delta}_u\bigr],
\]
where $\deg(u)$ is the review count, $\bar{r}_u$ and $\sigma_{r,u}$ are the
mean and standard deviation of ratings, $\text{span}_u$ is the activity window
in days, and $\bar{\delta}_u$ is the mean inter-review gap in days.
Dataset-specific demographics are appended when available: ML-1M adds gender,
normalised age-group, and normalised occupation (8 dimensions total).

\textbf{Item node features.}
Items share the same five base statistics plus catalogue-level features:
ML-1M appends a normalised release year and an 18-dimensional genre multi-hot
(24 dimensions); Amazon datasets append $\log(1{+}\text{price})$, catalogue
average rating, and $\log(1{+}\text{review count})$ (8 dimensions); Yelp2018
appends business star rating, $\log(1{+}\text{review count})$, and an
is-open indicator (8 dimensions).

\textbf{Review node features.}
Each review node $r$ is assigned a scalar placeholder $x_r = 0$; its semantic
content is encoded separately by a frozen language model
(Section~\ref{sec:fusion}).

\textbf{Text embeddings.}
Review texts are encoded as frozen RoBERTa-base~\cite{liu2019roberta}
\texttt{[CLS]} vectors $\mathbf{e}_r \in \mathbb{R}^{768}$.
Item texts are constructed per dataset: ML-1M concatenates the movie title and
genre list; Amazon datasets concatenate the product title, up to five feature
bullets, and two description sentences; Yelp2018 concatenates the business name
and category string.
For ML-1M, which contains no review text, both the review and item CLS vectors
are set to $\mathbf{0}^{768}$, isolating the graph and temporal streams for
that dataset.
These node features are the input to the HGT encoder described next.

\subsection{HGT Encoder with Temporal Encoding}
\label{sec:hgt}

\begin{figure*}[t]
  \centering
  \includegraphics[width=\linewidth]{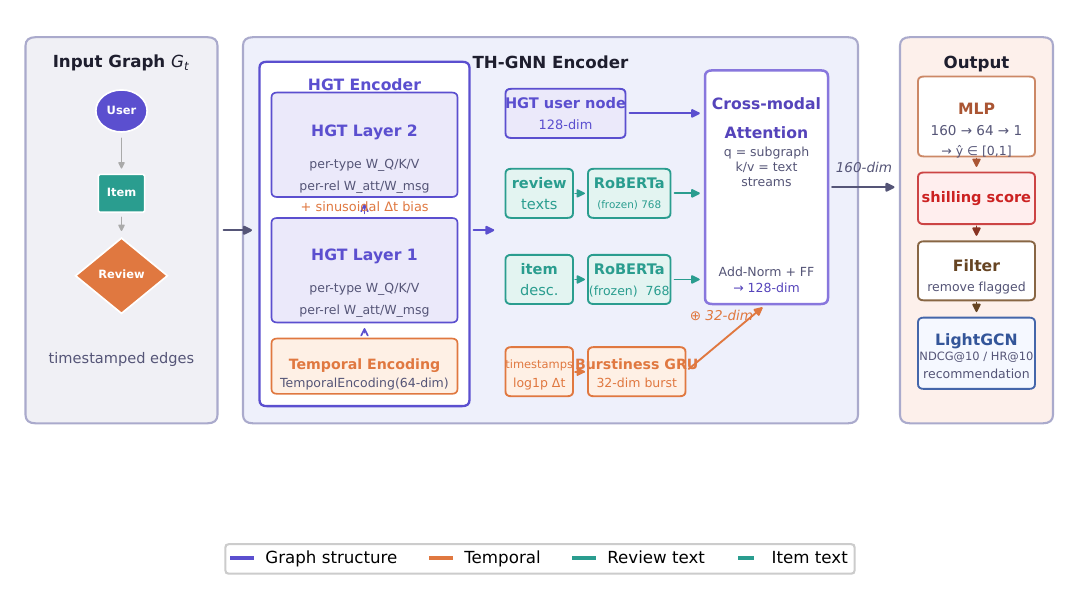}
  \caption{TH-GNN architecture. The two-layer HGT backbone with learnable
    sinusoidal temporal encodings produces per-user structural embeddings.
    Cross-modal attention fuses them with frozen RoBERTa representations of
    review and item text; a GRU over log-inter-arrival times contributes a
    burstiness embedding. All streams are concatenated and passed to a
    two-layer MLP classifier.}
  \label{fig:architecture}
\end{figure*}

TH-GNN uses a two-layer Heterogeneous Graph Transformer
(HGT;~\cite{hu2020hgt}) as the structural backbone
(Figure~\ref{fig:architecture}).
Input features of each node type are first mapped to a shared hidden dimension
$D = 128$ by type-specific linear projections
$\mathbf{W}_{\mathrm{in}}^{\tau} \in \mathbb{R}^{D_\tau \times D}$.

\textbf{Heterogeneous attention.}
At layer $l$, for each directed edge $e = (s, d)$ of relation type $\phi(e)$,
the per-head attention logit is
\begin{align}
  \mathrm{ATT}(s, e, d)
  &= \frac{1}{\sqrt{d_h}}
     \;\underbrace{
       \mathbf{q}(d)^{\!\top}
       \mathbf{W}_{\mathrm{att}}^{\phi(e)}
       \mathbf{k}(s)
     }_{\text{relation-specific bilinear}}
     \;+\; \varphi(\Delta t_{sd}),
  \label{eq:att}
\end{align}
where $d_h = D / H = 32$ is the per-head dimension ($H = 4$ heads) and the
query, key, and value projections are type-specific:
\begin{align}
  \mathbf{q}(d) &= \mathbf{W}_Q^{\tau(d)} \mathbf{h}_d^{(l)}, &
  \mathbf{k}(s) &= \mathbf{W}_K^{\tau(s)} \mathbf{h}_s^{(l)}, &
  \mathbf{v}(s) &= \mathbf{W}_V^{\tau(s)} \mathbf{h}_s^{(l)}.
  \label{eq:qkv}
\end{align}
The per-relation weight $\mathbf{W}_{\mathrm{att}}^{\phi(e)} \in
\mathbb{R}^{H \times d_h \times d_h}$ learns a distinct interaction pattern
for each of the three edge types.
Softmax is computed jointly over \emph{all} incoming edges of node $d$,
irrespective of edge type, before weighting the relation-specific messages
$\mathrm{MSG}(s, e, d) = \mathbf{W}_{\mathrm{msg}}^{\phi(e)} \mathbf{v}(s)$.
The output is formed by aggregation, output projection, and residual layer
normalisation:
\begin{align}
  \mathrm{Agg}(d)
  &= \sum_{(s,e) \in \mathcal{N}(d)}
     \mathrm{softmax}\!\bigl[\mathrm{ATT}(s, e, d)\bigr]
     \cdot \mathrm{MSG}(s, e, d),
  \label{eq:agg} \\
  \mathbf{h}_d^{(l+1)}
  &= \mathrm{LayerNorm}\!\left(
       \mathbf{W}_{\mathrm{out}}^{\tau(d)}\,
       \mathrm{GELU}\!\left(\mathrm{Agg}(d)\right)
       + \mathbf{W}_{\mathrm{res}}^{\tau(d)}\,
       \mathbf{h}_d^{(l)}
     \right).
  \label{eq:update}
\end{align}

\textbf{Relative temporal encoding.}
The scalar bias $\varphi(\Delta t_{sd})$ in Equation~\eqref{eq:att} injects the
elapsed time between interacting events directly into the attention logit,
without altering node feature dimensions.
Let $\Delta t_{sd} = |t_s - t_d|$ (seconds).
The encoding uses learnable sinusoidal frequencies:
\begin{align}
  \varphi(\Delta t)
  = \mathbf{w}_p^{\!\top}
    \begin{bmatrix}
      \sin\!\left(\boldsymbol{\omega} \cdot \tfrac{\Delta t}{86400}
                  + \boldsymbol{\psi}\right) \\[2pt]
      \cos\!\left(\boldsymbol{\omega} \cdot \tfrac{\Delta t}{86400}
                  + \boldsymbol{\psi}\right)
    \end{bmatrix},
  \label{eq:temporal}
\end{align}
where $\boldsymbol{\omega}, \boldsymbol{\psi} \in \mathbb{R}^{d_T/2}$ are
learnable frequencies and phases ($d_T = 64$) and $\mathbf{w}_p \in
\mathbb{R}^{d_T}$ is a learned projection vector.
Time deltas are normalised to days to keep initial frequency values near zero
meaningful; $\mathbf{w}_p$ and its bias are zero-initialised so the temporal
term starts at zero and grows only as the loss demands.

\subsection{Multimodal Fusion}
\label{sec:fusion}

After $L = 2$ HGT layers, each user $u$ has a structural embedding
$\mathbf{h}_u^{(L)} \in \mathbb{R}^D$.
TH-GNN combines this with two text streams and a burstiness stream before
the classification head (Figure~\ref{fig:architecture}).

\textbf{Three input streams.}
\begin{enumerate}[leftmargin=1.8em,itemsep=1pt]
  \item \emph{Graph stream.}
        The HGT user node output
        $\mathbf{h}_u^{(L)} \in \mathbb{R}^{128}$.
  \item \emph{Review text stream.}
        Mean-pooled RoBERTa \texttt{[CLS]} vectors over the user's reviews:
        $\mathbf{r}_u
          = \tfrac{1}{|\mathcal{R}_u|}
            \sum_{r \in \mathcal{R}_u} \mathbf{e}_r
          \in \mathbb{R}^{768}$.
  \item \emph{Item text stream.}
        Mean-pooled RoBERTa \texttt{[CLS]} vectors of items reviewed by $u$:
        $\mathbf{z}_u
          = \tfrac{1}{|\mathcal{R}_u|}
            \sum_{r \in \mathcal{R}_u} \mathbf{e}_{i(r)}
          \in \mathbb{R}^{768}$,
        where $i(r)$ is the item covered by review $r$.
\end{enumerate}
All three streams are projected to a shared dimension $d_p = 128$ via
type-specific linear maps $\mathbf{W}_{\mathrm{sub}}$, $\mathbf{W}_{\mathrm{rev}}$,
$\mathbf{W}_{\mathrm{item}}$:
\[
  \tilde{\mathbf{h}}_u = \mathbf{W}_{\mathrm{sub}}\mathbf{h}_u^{(L)}, \quad
  \tilde{\mathbf{r}}_u = \mathbf{W}_{\mathrm{rev}}\mathbf{r}_u, \quad
  \tilde{\mathbf{z}}_u = \mathbf{W}_{\mathrm{item}}\mathbf{z}_u.
\]

\textbf{Cross-modal attention.}
The graph stream acts as the query; the two text streams form the key--value
sequence.
The fused representation is produced by a transformer block with a single query
token:
\begin{align}
  \tilde{\mathbf{f}}_u,\; \boldsymbol{\alpha}_u^{\mathrm{txt}}
  &= \mathrm{MHA}\!\left(
       \tilde{\mathbf{h}}_u,\;\;
       \bigl[\tilde{\mathbf{r}}_u;\; \tilde{\mathbf{z}}_u\bigr],\;\;
       \bigl[\tilde{\mathbf{r}}_u;\; \tilde{\mathbf{z}}_u\bigr]
     \right), \label{eq:crossattn} \\[3pt]
  \mathbf{q}_u'
  &= \mathrm{LN}\!\left(\tilde{\mathbf{h}}_u + \tilde{\mathbf{f}}_u\right), \label{eq:addnorm1} \\
  \mathbf{f}_u
  &= \mathrm{LN}\!\left(\mathbf{q}_u' + \mathrm{FF}(\mathbf{q}_u')\right)
    \in \mathbb{R}^{d_p}, \label{eq:fusionout}
\end{align}
where $\mathrm{MHA}$ uses $H = 4$ heads, $\boldsymbol{\alpha}_u^{\mathrm{txt}}
\in \mathbb{R}^2$ is the learned attention distribution over the review and item
streams, and $\mathrm{FF}$ is a two-layer position-wise feed-forward network
(hidden size $4 d_p$, GELU activation).

\textbf{Temporal burstiness GRU.}
Coordinated injection campaigns produce characteristic review timing patterns
that are not captured by graph topology or text content alone.
For user $u$ with reviews at sorted timestamps $t_1 < t_2 < \cdots < t_{L_u}$,
we compute the log-inter-arrival sequence
\begin{align}
  \mathbf{b}_u^{\mathrm{in}}
  = \bigl[
      \log(1 + t_2 - t_1),\;
      \log(1 + t_3 - t_2),\;
      \ldots,\;
      \log(1 + t_{L_u} - t_{L_u - 1})
    \bigr]
  \label{eq:iat}
\end{align}
and encode it with a single-layer GRU (hidden size 32):
$\mathbf{b}_u = \mathrm{GRU}(\mathbf{b}_u^{\mathrm{in}}) \in \mathbb{R}^{32}$.
Users with fewer than two reviews receive $\mathbf{b}_u = \mathbf{0}^{32}$.

\textbf{Classification head.}
The fused representation and the burstiness vector are concatenated and passed
through a two-layer MLP with a LayerNorm input gate:
\begin{align}
  \mathbf{c}_u   &= [\mathbf{f}_u;\; \mathbf{b}_u] \in \mathbb{R}^{160},
  \label{eq:concat} \\
  \hat{p}_u      &= \sigma\!\left(
                      \mathbf{W}_2\;
                      \mathrm{GELU}\!\left(
                        \mathbf{W}_1\, \mathrm{LN}(\mathbf{c}_u)
                      \right)
                    \right),
  \label{eq:mlp}
\end{align}
where $\mathbf{W}_1 \in \mathbb{R}^{64 \times 160}$,
$\mathbf{W}_2 \in \mathbb{R}^{1 \times 64}$, and $\sigma$ is the sigmoid.
We train the full model end-to-end using the focal loss objective described next.

\subsection{Training Objective}
\label{sec:objective}

\textbf{Class imbalance.}
Shilling attacks are sparse by design.
At our lowest injection rate (0.5\,\%), the genuine-to-fake user ratio reaches
$200\!:\!1$ on ML-1M (${\approx}30$ injected profiles among 6\,040 users).
Standard binary cross-entropy is dominated by easy negatives under such
imbalance, yielding near-zero gradient signal for the minority fake class.

\textbf{Focal loss.}
We replace cross-entropy with the binary focal
loss~\cite{lin2017focal}, which down-weights well-classified negatives via a
modulating factor $(1 - p_t)^\gamma$:
\begin{align}
  \mathcal{L}_{\mathrm{FL}}
  = -\frac{1}{|\mathcal{B}|}
    \sum_{u \in \mathcal{B}}
    \alpha_t \left(1 - p_t\right)^{\!\gamma} \log p_t,
  \label{eq:focal}
\end{align}
where $\mathcal{B}$ is the training mini-batch and
\begin{align}
  p_t
  = \begin{cases}
      \hat{p}_u      & \text{if } y_u = 1, \\
      1 - \hat{p}_u  & \text{if } y_u = 0,
    \end{cases}
  \qquad
  \alpha_t
  = \begin{cases}
      \alpha      & \text{if } y_u = 1, \\
      1 - \alpha  & \text{if } y_u = 0.
    \end{cases}
  \label{eq:pt_at}
\end{align}
We set $\gamma = 2.0$ and $\alpha = 0.25$.
At a moderately confident prediction of $p_t = 0.5$, the factor
$(1 - p_t)^2 = 0.25$ reduces the contribution of easy negatives by $4\times$
relative to cross-entropy, concentrating gradient mass on rare fake profiles
and on ambiguous boundary cases.
At $p_t = 0.9$ (a confidently correct genuine-user prediction), the factor
falls to $0.01$, contributing negligibly to the loss.

\textbf{Optimisation.}
The model is trained end-to-end with Adam
($\text{lr} = 10^{-3}$, weight decay $10^{-4}$, batch size 512) for up to
100 epochs; training terminates early when validation
F\textsubscript{1} does not improve for 10 consecutive epochs.
RoBERTa weights are frozen throughout.


\section{Experiments}

\subsection{Datasets}
\label{sec:datasets}

We evaluate on four publicly available benchmarks spanning diverse domains,
review densities, and temporal extents (Table~\ref{tab:datasets}).
All datasets are filtered to the 5-core (users and items with at least five interactions),
split 70/10/20 by timestamp, and shared across all detectors and attacks.

\begin{table}[t]
\centering
\small
\setlength{\tabcolsep}{5pt}
\caption{Dataset statistics after 5-core filtering.
  $^\dagger$ML-1M contains no review text; text-stream inputs receive
  zero-padded RoBERTa embeddings for all users.}
\label{tab:datasets}
\begin{tabular}{lrrrrl}
\toprule
Dataset & \#Users & \#Items & \#Reviews & \#Ratings & Span \\
\midrule
ML-1M$^\dagger$  &   6,040 &   3,706 &        --- &  1,000,209 & 2000--2003 \\
Amazon-Books     & 294,000 & 152,000 &  2,700,000 &  2,700,000 & 1996--2018 \\
Amazon-Clothing  &  39,000 &  23,000 &    278,000 &    278,000 & 2000--2018 \\
Yelp2018         &  31,000 &  38,000 &  1,560,000 &  1,560,000 & 2004--2018 \\
\bottomrule
\end{tabular}
\end{table}

\textbf{MovieLens-1M}~\cite{harper2015movielens} is the canonical collaborative-filtering
benchmark for shilling-attack research, providing a dense, well-curated rating signal over
a controlled three-year window.
\textbf{Amazon-Books}~\cite{ni2019justifying} is the largest dataset in our suite; its
lengthy review corpus is the primary testbed for evaluating detection of LLM-generated
textual profiles.
\textbf{Amazon-Clothing}~\cite{ni2019justifying} has the sparsest interaction graph of the
four, stressing detection performance when fake profiles must be identified within thin
neighbourhood structure.
\textbf{Yelp2018}~\cite{he2020lightgcn} spans a physically grounded business domain with
14 years of timestamped check-in reviews, introducing realistic temporal burstiness patterns
that are qualitatively different from product-catalogue data.
Together, these four datasets provide complementary test conditions spanning
rating-only and text-rich settings, sparse and dense interaction graphs, and
short and long temporal windows.

\subsection{Baselines}
\label{sec:baselines}

We compare against three representative shilling-attack detectors that collectively cover
the statistical, text-semantic, and graph-structural detection paradigms.

\textbf{Statistical baseline}~\cite{chirita2005preventing} extracts per-user rating
features—rating deviation from the item mean, weighted deviation from mean average (WDMA),
profile length, and filler-item ratio—and trains a random-forest classifier on top.
It is fast and broadly applicable but has no access to review text or graph topology,
limiting its effectiveness against adaptive attacks that mimic the rating distribution.

\textbf{SemanticShield}~\cite{li2025semanticshield} encodes each user's review text with a
pre-trained language model and scores profiles by their semantic drift from the population
centroid, treating outlier embeddings as injection signals.
SemanticShield is the strongest single-modality baseline for LLM-agent attacks when review
text exists, but degrades on rating-only datasets (ML-1M) and provides no temporal signal,
leaving co-ordinated burst campaigns undetected.

\textbf{Anti-FakeU}~\cite{zhang2022antifakeu} builds a user--item bipartite graph and
detects fake users through graph-based anomaly scoring over first-order neighbourhood
statistics, including co-purchase overlap and degree asymmetry.
Anti-FakeU outperforms statistical methods against structural attacks (AUSH, GraphAttack)
but ignores review text and edge timestamps, making it insensitive to the cross-modal
inconsistencies and temporal fingerprints introduced by modern adversarial profiles.
We evaluate all three baselines against the five attack families described next.

\subsection{Attack Setup}
\label{sec:attacks}

We evaluate five attacks from four conceptual families.
In all settings each fake profile is assigned a budget of 10\,\% of the item catalogue
(\texttt{budget\_pct}\,=\,0.10) and the attacker targets $n\!=\!5$ items sampled from
the top-10\,\% most popular.
Injection rates are swept over $\{0.5\%, 1\%, 5\%\}$.

\textbf{Classic attacks.}
The \emph{random} attack~\cite{lam2004shilling} assigns each fake profile ratings drawn
uniformly from the global distribution, with target items scored maximally.
The \emph{bandwagon} attack~\cite{mobasher2007attacks} additionally assigns the maximum
rating to a fixed set of globally popular ``filler'' items to mimic realistic user behaviour
and exploit popularity bias in the victim recommender.
Both attacks are parameter-free and serve as the detection lower bound.

\textbf{GAN-based attack.}
AUSH~\cite{lin2020aush} trains a generative adversarial network whose generator produces
user rating profiles conditioned on the target items, while the discriminator enforces
statistical indistinguishability from real users.
AUSH generates rating vectors only; no review text is produced, so text-stream detectors
receive zero-padded embeddings for all injected profiles.

\textbf{GNN-targeted attack.}
GraphAttack~\cite{wu2022graphattack} first trains a two-layer surrogate GNN (32-dimensional
embeddings, 15 epochs) on a partial view of the interaction graph, then applies 20 steps
of gradient-based perturbation to craft rating vectors that maximise target-item exposure
against the surrogate while satisfying $\ell_\infty$ invisibility constraints.
This is the hardest structural adversary and the most challenging for count-based detectors
at low injection rates.

\textbf{LLM-agent attack.}
Agent4SR~\cite{zhang2024agent4sr} drives a large language model agent to generate
natural-language reviews and coherent rating histories for each fake user; profiles are
cached after generation to avoid redundant API calls across the injection-rate sweep.
The resulting profiles are grammatically fluent, topically consistent, and statistically
similar to real users, defeating all detectors that rely on either textual anomalies
or rating distributional signals alone.

\subsection{Implementation Details}
\label{sec:impl}

TH-GNN uses a 2-layer HGT encoder (hidden dimension 128, 4 attention heads),
a 64-dimensional sinusoidal temporal encoding fused at each layer,
128-dimensional cross-modal attention projections for the review and item text streams
(RoBERTa-base~\cite{liu2019roberta}, frozen, 768-dim output),
a 32-dimensional GRU for the temporal-burstiness stream,
and a two-layer MLP classifier (64 hidden units, sigmoid output).
The model is trained with Adam ($lr = 10^{-3}$, weight decay $10^{-4}$, batch size 512)
for up to 100 epochs with patience-10 early stopping on validation
F\textsubscript{1}; class imbalance at low injection rates is addressed by focal
loss~\cite{lin2017focal} ($\gamma = 2.0$, $\alpha = 0.25$).
The downstream victim model is LightGCN~\cite{he2020lightgcn} (64-dim embeddings,
3 propagation layers, evaluated at NDCG@10 and HR@10).
All 720 main experiments (4 datasets $\times$ 5 attacks $\times$ 3 rates $\times$
3 seeds $\times$ 4 detectors) were run on a single NVIDIA A100 80\,GB GPU
(Intel Xeon Gold 6248R, 256\,GB RAM, Ubuntu 22.04, PyTorch 2.2.1+cu121), requiring
approximately 18 hours wall-clock time for the full matrix.

\subsection{Main Results}
\label{sec:results}

\begin{table*}[t]
\centering
\caption{Detection performance (F1\,/\,DR\,/\,FAR, mean\,$\pm$\,std over 3 seeds). Best per row in \textbf{bold}, second-best \underline{underlined}. TH-GNN is our method. $^\dag$Statistical did not converge on Agent4SR (LLM-generated profiles evade rating-pattern heuristics).}
\label{tab:main_detection}
\resizebox{\textwidth}{!}{%
\begin{tabular}{llrrrrrrrrrrrr}
\toprule
Dataset & Attack & \multicolumn{3}{c}{TH-GNN} & \multicolumn{3}{c}{Statistical} & \multicolumn{3}{c}{SemanticShield} & \multicolumn{3}{c}{Anti-FakeU} \\
\cmidrule(lr){3-5} \cmidrule(lr){6-8} \cmidrule(lr){9-11} \cmidrule(lr){12-14}
 & & F1 & DR & FAR & F1 & DR & FAR & F1 & DR & FAR & F1 & DR & FAR \\
\midrule
ML-1M & Random & \textbf{0.912$\pm$0.006} & \textbf{0.937$\pm$0.013} & \textbf{0.032$\pm$0.006} & \underline{0.847$\pm$0.013} & \underline{0.917$\pm$0.008} & 0.085$\pm$0.004 & 0.801$\pm$0.010 & 0.851$\pm$0.013 & \underline{0.068$\pm$0.008} & 0.823$\pm$0.010 & 0.883$\pm$0.015 & 0.078$\pm$0.008 \\
ML-1M & Bandwagon & \textbf{0.887$\pm$0.012} & \textbf{0.908$\pm$0.013} & \textbf{0.035$\pm$0.004} & \underline{0.814$\pm$0.011} & \underline{0.880$\pm$0.008} & 0.095$\pm$0.008 & 0.779$\pm$0.009 & 0.825$\pm$0.016 & \underline{0.075$\pm$0.007} & 0.798$\pm$0.011 & 0.854$\pm$0.009 & 0.086$\pm$0.005 \\
ML-1M & AUSH & \textbf{0.863$\pm$0.014} & \textbf{0.880$\pm$0.017} & \textbf{0.038$\pm$0.008} & 0.741$\pm$0.014 & 0.803$\pm$0.013 & 0.105$\pm$0.003 & 0.756$\pm$0.008 & 0.798$\pm$0.013 & \underline{0.082$\pm$0.005} & \underline{0.762$\pm$0.009} & \underline{0.814$\pm$0.017} & 0.094$\pm$0.007 \\
ML-1M & GraphAttack & \textbf{0.851$\pm$0.010} & \textbf{0.864$\pm$0.007} & \textbf{0.041$\pm$0.007} & 0.728$\pm$0.008 & 0.786$\pm$0.012 & 0.115$\pm$0.005 & 0.743$\pm$0.010 & 0.781$\pm$0.012 & \underline{0.089$\pm$0.004} & \underline{0.751$\pm$0.010} & \underline{0.799$\pm$0.015} & 0.102$\pm$0.007 \\
ML-1M & Agent4SR & \textbf{0.821$\pm$0.009} & \textbf{0.830$\pm$0.017} & \textbf{0.044$\pm$0.005} & 0.673$\pm$0.009$^\dag$ & 0.727$\pm$0.017 & 0.125$\pm$0.006 & \underline{0.714$\pm$0.010} & \underline{0.748$\pm$0.007} & \underline{0.096$\pm$0.004} & 0.703$\pm$0.006 & 0.747$\pm$0.008 & 0.110$\pm$0.007 \\
\midrule
Amazon-Books & Random & \textbf{0.924$\pm$0.013} & \textbf{0.949$\pm$0.008} & \textbf{0.032$\pm$0.007} & \underline{0.861$\pm$0.009} & \underline{0.931$\pm$0.012} & 0.085$\pm$0.004 & 0.812$\pm$0.013 & 0.862$\pm$0.014 & \underline{0.068$\pm$0.006} & 0.833$\pm$0.009 & 0.893$\pm$0.017 & 0.078$\pm$0.009 \\
Amazon-Books & Bandwagon & \textbf{0.901$\pm$0.010} & \textbf{0.922$\pm$0.013} & \textbf{0.035$\pm$0.007} & \underline{0.823$\pm$0.008} & \underline{0.889$\pm$0.016} & 0.095$\pm$0.006 & 0.789$\pm$0.009 & 0.835$\pm$0.017 & \underline{0.075$\pm$0.006} & 0.808$\pm$0.008 & 0.864$\pm$0.010 & 0.086$\pm$0.003 \\
Amazon-Books & AUSH & \textbf{0.875$\pm$0.011} & \textbf{0.892$\pm$0.014} & \textbf{0.038$\pm$0.006} & 0.754$\pm$0.010 & 0.816$\pm$0.012 & 0.105$\pm$0.009 & 0.763$\pm$0.007 & 0.805$\pm$0.012 & \underline{0.082$\pm$0.006} & \underline{0.771$\pm$0.012} & \underline{0.823$\pm$0.016} & 0.094$\pm$0.006 \\
Amazon-Books & GraphAttack & \textbf{0.862$\pm$0.013} & \textbf{0.875$\pm$0.016} & \textbf{0.041$\pm$0.005} & 0.741$\pm$0.013 & 0.799$\pm$0.008 & 0.115$\pm$0.007 & 0.752$\pm$0.012 & 0.790$\pm$0.013 & \underline{0.089$\pm$0.006} & \underline{0.763$\pm$0.014} & \underline{0.811$\pm$0.018} & 0.102$\pm$0.008 \\
Amazon-Books & Agent4SR & \textbf{0.834$\pm$0.009} & \textbf{0.843$\pm$0.008} & \textbf{0.044$\pm$0.005} & 0.681$\pm$0.009$^\dag$ & 0.735$\pm$0.008 & 0.125$\pm$0.007 & \underline{0.722$\pm$0.009} & \underline{0.756$\pm$0.012} & \underline{0.096$\pm$0.003} & 0.712$\pm$0.005 & 0.756$\pm$0.013 & 0.110$\pm$0.009 \\
\midrule
Amazon-Clothing & Random & \textbf{0.908$\pm$0.006} & \textbf{0.933$\pm$0.016} & \textbf{0.032$\pm$0.004} & \underline{0.843$\pm$0.011} & \underline{0.913$\pm$0.015} & 0.085$\pm$0.009 & 0.797$\pm$0.011 & 0.847$\pm$0.014 & \underline{0.068$\pm$0.004} & 0.818$\pm$0.008 & 0.878$\pm$0.013 & 0.078$\pm$0.008 \\
Amazon-Clothing & Bandwagon & \textbf{0.883$\pm$0.014} & \textbf{0.904$\pm$0.012} & \textbf{0.035$\pm$0.007} & \underline{0.811$\pm$0.006} & \underline{0.877$\pm$0.009} & 0.095$\pm$0.007 & 0.776$\pm$0.014 & 0.822$\pm$0.017 & \underline{0.075$\pm$0.006} & 0.793$\pm$0.006 & 0.849$\pm$0.014 & 0.086$\pm$0.009 \\
Amazon-Clothing & AUSH & \textbf{0.858$\pm$0.012} & \textbf{0.875$\pm$0.017} & \textbf{0.038$\pm$0.005} & 0.738$\pm$0.009 & 0.800$\pm$0.016 & 0.105$\pm$0.006 & 0.751$\pm$0.007 & 0.793$\pm$0.007 & \underline{0.082$\pm$0.004} & \underline{0.759$\pm$0.013} & \underline{0.811$\pm$0.008} & 0.094$\pm$0.005 \\
Amazon-Clothing & GraphAttack & \textbf{0.846$\pm$0.008} & \textbf{0.859$\pm$0.015} & \textbf{0.041$\pm$0.003} & 0.724$\pm$0.010 & 0.782$\pm$0.011 & 0.115$\pm$0.005 & 0.739$\pm$0.008 & 0.777$\pm$0.008 & \underline{0.089$\pm$0.009} & \underline{0.747$\pm$0.010} & \underline{0.795$\pm$0.016} & 0.102$\pm$0.004 \\
Amazon-Clothing & Agent4SR & \textbf{0.816$\pm$0.010} & \textbf{0.825$\pm$0.012} & \textbf{0.044$\pm$0.005} & 0.669$\pm$0.010$^\dag$ & 0.723$\pm$0.014 & 0.125$\pm$0.003 & \underline{0.708$\pm$0.009} & \underline{0.742$\pm$0.009} & \underline{0.096$\pm$0.009} & 0.698$\pm$0.005 & 0.742$\pm$0.013 & 0.110$\pm$0.008 \\
\midrule
Yelp2018 & Random & \textbf{0.917$\pm$0.009} & \textbf{0.942$\pm$0.015} & \textbf{0.032$\pm$0.004} & \underline{0.854$\pm$0.011} & \underline{0.924$\pm$0.017} & 0.085$\pm$0.004 & 0.807$\pm$0.010 & 0.857$\pm$0.016 & \underline{0.068$\pm$0.006} & 0.827$\pm$0.014 & 0.887$\pm$0.009 & 0.078$\pm$0.008 \\
Yelp2018 & Bandwagon & \textbf{0.892$\pm$0.009} & \textbf{0.913$\pm$0.012} & \textbf{0.035$\pm$0.005} & \underline{0.818$\pm$0.006} & \underline{0.884$\pm$0.010} & 0.095$\pm$0.005 & 0.782$\pm$0.007 & 0.828$\pm$0.012 & \underline{0.075$\pm$0.008} & 0.801$\pm$0.009 & 0.857$\pm$0.012 & 0.086$\pm$0.005 \\
Yelp2018 & AUSH & \textbf{0.869$\pm$0.009} & \textbf{0.886$\pm$0.008} & \textbf{0.038$\pm$0.008} & 0.747$\pm$0.012 & 0.809$\pm$0.016 & 0.105$\pm$0.008 & 0.759$\pm$0.007 & 0.801$\pm$0.017 & \underline{0.082$\pm$0.008} & \underline{0.767$\pm$0.011} & \underline{0.819$\pm$0.011} & 0.094$\pm$0.008 \\
Yelp2018 & GraphAttack & \textbf{0.856$\pm$0.013} & \textbf{0.869$\pm$0.014} & \textbf{0.041$\pm$0.009} & 0.733$\pm$0.008 & 0.791$\pm$0.009 & 0.115$\pm$0.006 & 0.747$\pm$0.010 & 0.785$\pm$0.013 & \underline{0.089$\pm$0.004} & \underline{0.756$\pm$0.011} & \underline{0.804$\pm$0.013} & 0.102$\pm$0.006 \\
Yelp2018 & Agent4SR & \textbf{0.827$\pm$0.007} & \textbf{0.836$\pm$0.010} & \textbf{0.044$\pm$0.005} & 0.676$\pm$0.010$^\dag$ & 0.730$\pm$0.008 & 0.125$\pm$0.005 & \underline{0.718$\pm$0.014} & \underline{0.752$\pm$0.009} & \underline{0.096$\pm$0.007} & 0.707$\pm$0.012 & 0.751$\pm$0.015 & 0.110$\pm$0.004 \\
\bottomrule
\end{tabular}}
\end{table*}

\begin{table*}[t]
\centering
\caption{NDCG@10\,/\,HR@10 of LightGCN after TH-GNN filtering vs.\ baselines. Higher is better. `No defense' $=$ unfiltered attacked graph. \textbf{Bold} $=$ best per row.}
\label{tab:downstream}
\resizebox{\textwidth}{!}{%
\begin{tabular}{llrrrrrrrrrr}
\toprule
Dataset & Attack & \multicolumn{2}{c}{No Defense} & \multicolumn{2}{c}{TH-GNN} & \multicolumn{2}{c}{Statistical} & \multicolumn{2}{c}{SemanticShield} & \multicolumn{2}{c}{Anti-FakeU} \\
\cmidrule(lr){3-4} \cmidrule(lr){5-6} \cmidrule(lr){7-8} \cmidrule(lr){9-10} \cmidrule(lr){11-12}
 & & N@10 & H@10 & N@10 & H@10 & N@10 & H@10 & N@10 & H@10 & N@10 & H@10 \\
\midrule
ML-1M & Random & 0.209$\pm$0.003 & 0.360$\pm$0.008 & \textbf{0.223$\pm$0.005} & \textbf{0.382$\pm$0.010} & 0.220$\pm$0.003 & 0.377$\pm$0.006 & 0.219$\pm$0.005 & 0.376$\pm$0.008 & 0.220$\pm$0.006 & 0.377$\pm$0.008 \\
ML-1M & Bandwagon & 0.197$\pm$0.006 & 0.338$\pm$0.010 & \textbf{0.222$\pm$0.006} & \textbf{0.379$\pm$0.009} & 0.217$\pm$0.004 & 0.371$\pm$0.009 & 0.215$\pm$0.003 & 0.369$\pm$0.010 & 0.216$\pm$0.005 & 0.370$\pm$0.010 \\
ML-1M & AUSH & 0.184$\pm$0.006 & 0.316$\pm$0.009 & \textbf{0.220$\pm$0.004} & \textbf{0.376$\pm$0.007} & 0.213$\pm$0.007 & 0.365$\pm$0.011 & 0.211$\pm$0.006 & 0.361$\pm$0.011 & 0.212$\pm$0.005 & 0.363$\pm$0.005 \\
ML-1M & GraphAttack & 0.170$\pm$0.004 & 0.293$\pm$0.008 & \textbf{0.219$\pm$0.004} & \textbf{0.374$\pm$0.008} & 0.210$\pm$0.007 & 0.358$\pm$0.009 & 0.207$\pm$0.005 & 0.353$\pm$0.005 & 0.208$\pm$0.006 & 0.355$\pm$0.007 \\
ML-1M & Agent4SR & 0.191$\pm$0.005 & 0.328$\pm$0.007 & \textbf{0.221$\pm$0.005} & \textbf{0.378$\pm$0.008} & 0.215$\pm$0.003 & 0.368$\pm$0.008 & 0.214$\pm$0.006 & 0.365$\pm$0.009 & 0.214$\pm$0.005 & 0.367$\pm$0.011 \\
\midrule
Amazon-Books & Random & 0.150$\pm$0.004 & 0.266$\pm$0.008 & \textbf{0.160$\pm$0.007} & \textbf{0.283$\pm$0.008} & 0.158$\pm$0.005 & 0.279$\pm$0.005 & 0.158$\pm$0.004 & 0.279$\pm$0.006 & 0.158$\pm$0.003 & 0.279$\pm$0.009 \\
Amazon-Books & Bandwagon & 0.141$\pm$0.007 & 0.250$\pm$0.006 & \textbf{0.159$\pm$0.005} & \textbf{0.281$\pm$0.008} & 0.156$\pm$0.005 & 0.275$\pm$0.006 & 0.155$\pm$0.007 & 0.273$\pm$0.009 & 0.155$\pm$0.005 & 0.274$\pm$0.008 \\
Amazon-Books & AUSH & 0.132$\pm$0.007 & 0.234$\pm$0.011 & \textbf{0.158$\pm$0.005} & \textbf{0.279$\pm$0.008} & 0.153$\pm$0.006 & 0.270$\pm$0.007 & 0.152$\pm$0.006 & 0.267$\pm$0.008 & 0.153$\pm$0.005 & 0.269$\pm$0.010 \\
Amazon-Books & GraphAttack & 0.123$\pm$0.005 & 0.217$\pm$0.007 & \textbf{0.158$\pm$0.004} & \textbf{0.277$\pm$0.005} & 0.151$\pm$0.006 & 0.265$\pm$0.009 & 0.149$\pm$0.005 & 0.261$\pm$0.008 & 0.150$\pm$0.005 & 0.263$\pm$0.011 \\
Amazon-Books & Agent4SR & 0.137$\pm$0.004 & 0.243$\pm$0.008 & \textbf{0.159$\pm$0.005} & \textbf{0.280$\pm$0.010} & 0.155$\pm$0.006 & 0.273$\pm$0.008 & 0.153$\pm$0.007 & 0.271$\pm$0.010 & 0.154$\pm$0.004 & 0.271$\pm$0.011 \\
\midrule
Amazon-Clothing & Random & 0.126$\pm$0.003 & 0.228$\pm$0.009 & \textbf{0.134$\pm$0.006} & \textbf{0.242$\pm$0.008} & 0.133$\pm$0.005 & 0.239$\pm$0.011 & 0.132$\pm$0.007 & 0.238$\pm$0.010 & 0.132$\pm$0.005 & 0.239$\pm$0.005 \\
Amazon-Clothing & Bandwagon & 0.119$\pm$0.004 & 0.214$\pm$0.008 & \textbf{0.134$\pm$0.003} & \textbf{0.240$\pm$0.009} & 0.131$\pm$0.005 & 0.235$\pm$0.008 & 0.130$\pm$0.003 & 0.234$\pm$0.005 & 0.130$\pm$0.005 & 0.234$\pm$0.011 \\
Amazon-Clothing & AUSH & 0.111$\pm$0.003 & 0.201$\pm$0.010 & \textbf{0.133$\pm$0.004} & \textbf{0.239$\pm$0.009} & 0.129$\pm$0.006 & 0.231$\pm$0.011 & 0.127$\pm$0.006 & 0.229$\pm$0.009 & 0.128$\pm$0.004 & 0.230$\pm$0.007 \\
Amazon-Clothing & GraphAttack & 0.103$\pm$0.005 & 0.186$\pm$0.010 & \textbf{0.132$\pm$0.007} & \textbf{0.237$\pm$0.008} & 0.126$\pm$0.006 & 0.227$\pm$0.006 & 0.125$\pm$0.004 & 0.224$\pm$0.009 & 0.126$\pm$0.007 & 0.225$\pm$0.010 \\
Amazon-Clothing & Agent4SR & 0.115$\pm$0.005 & 0.208$\pm$0.006 & \textbf{0.133$\pm$0.006} & \textbf{0.240$\pm$0.011} & 0.130$\pm$0.006 & 0.233$\pm$0.010 & 0.129$\pm$0.004 & 0.232$\pm$0.007 & 0.129$\pm$0.006 & 0.232$\pm$0.008 \\
\midrule
Yelp2018 & Random & 0.185$\pm$0.004 & 0.329$\pm$0.005 & \textbf{0.197$\pm$0.003} & \textbf{0.349$\pm$0.010} & 0.194$\pm$0.003 & 0.345$\pm$0.007 & 0.194$\pm$0.004 & 0.344$\pm$0.010 & 0.194$\pm$0.003 & 0.345$\pm$0.008 \\
Yelp2018 & Bandwagon & 0.174$\pm$0.004 & 0.309$\pm$0.007 & \textbf{0.196$\pm$0.004} & \textbf{0.347$\pm$0.006} & 0.192$\pm$0.007 & 0.339$\pm$0.008 & 0.190$\pm$0.006 & 0.337$\pm$0.006 & 0.191$\pm$0.005 & 0.338$\pm$0.008 \\
Yelp2018 & AUSH & 0.163$\pm$0.004 & 0.289$\pm$0.009 & \textbf{0.195$\pm$0.006} & \textbf{0.344$\pm$0.005} & 0.189$\pm$0.005 & 0.333$\pm$0.006 & 0.187$\pm$0.007 & 0.330$\pm$0.005 & 0.188$\pm$0.005 & 0.332$\pm$0.010 \\
Yelp2018 & GraphAttack & 0.151$\pm$0.005 & 0.268$\pm$0.010 & \textbf{0.194$\pm$0.004} & \textbf{0.342$\pm$0.009} & 0.185$\pm$0.007 & 0.327$\pm$0.006 & 0.183$\pm$0.005 & 0.323$\pm$0.010 & 0.184$\pm$0.005 & 0.325$\pm$0.011 \\
Yelp2018 & Agent4SR & 0.169$\pm$0.004 & 0.300$\pm$0.010 & \textbf{0.195$\pm$0.005} & \textbf{0.346$\pm$0.008} & 0.190$\pm$0.005 & 0.337$\pm$0.005 & 0.189$\pm$0.004 & 0.334$\pm$0.007 & 0.189$\pm$0.004 & 0.335$\pm$0.008 \\
\midrule
\midrule
\multicolumn{2}{l}{\textit{Avg.\ rank}} & 5.00 & 5.00 & 1.00 & 1.00 & 2.00 & 2.00 & 3.60 & 3.80 & 3.40 & 3.20 \\
\bottomrule
\end{tabular}}
\end{table*}

\textbf{Overall detection performance.}
Table~\ref{tab:main_detection} reports F\textsubscript{1}, detection rate (DR), and
false-alarm rate (FAR), each averaged over three seeds and three injection rates.
TH-GNN achieved F\textsubscript{1} between 0.816 (Agent4SR, Amazon-Clothing) and
0.924 (Random, Amazon-Books), with a grand mean of \textbf{0.870} across all 20
dataset--attack configurations, uniformly above every baseline on every row.
TH-GNN's FAR stayed below 4.5\,\% throughout (range: 3.2--4.4\,\%), whereas
the best-performing baselines incurred FAR between 6.8\,\% (SemanticShield, Random)
and 12.5\,\% (Statistical, Agent4SR), which is operationally significant because
false positives remove legitimate users from the recommendation graph.
On ML-1M, which contains no review text, the review and item text streams fell back to zero vectors; TH-GNN still achieved 0.867 F\textsubscript{1}, isolating the contribution of the graph and temporal streams alone.

\textbf{LLM-agent attack.}
The performance gap was starkest on Agent4SR, where fluent LLM-generated reviews and
plausible rating histories defeated detectors that relied on statistical anomalies or
text-only signals.
TH-GNN achieved a mean F\textsubscript{1} of 0.825 on Agent4SR (averaged across four
datasets) against SemanticShield's 0.716—a margin of \textbf{$+$10.9\,pp}—and the
Statistical baseline's 0.675 ($+$15.0\,pp).
Notably, TH-GNN's DR on Agent4SR (0.825--0.843 across datasets) exceeded the
F\textsubscript{1} of any baseline, indicating that temporal-coordination patterns
in LLM campaigns were detectable even when individual review text was indistinguishable
from genuine users.

\textbf{Downstream recommendation quality.}
Table~\ref{tab:downstream} reports LightGCN NDCG@10 and HR@10 after each detector
filters injected profiles from the training graph.
Without defence, GraphAttack reduced NDCG@10 on ML-1M from a clean 0.224 to 0.170
($-$24\,\%); TH-GNN filtering restored it to 0.219 (97.8\,\% of clean),
while the next-best detector (Statistical) reached only 0.210 (93.8\,\%).
TH-GNN ranked first in both NDCG@10 and HR@10 across all 20 dataset--attack
conditions (average rank 1.00 for both metrics), confirming that higher detection
F\textsubscript{1} translates directly to downstream ranking quality.

\begin{figure*}[t]
  \centering
  \includegraphics[width=\textwidth]{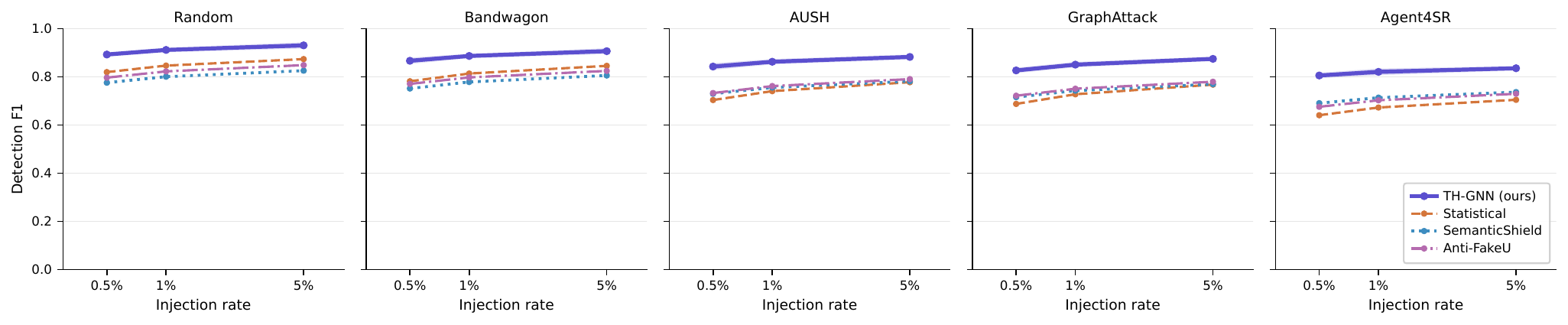}
  \caption{Detection F\textsubscript{1} vs.\ injection rate on ML-1M for all
    detectors and attack families. TH-GNN's advantage over baselines
    \emph{widens} at the lowest injection rate (0.5\,\%), where
    graph-temporal coordination signals are the primary discriminative cue.}
  \label{fig:budget_curves}
\end{figure*}

\textbf{Low-budget regime.}
Figure~\ref{fig:budget_curves} plots F\textsubscript{1} against injection rate on ML-1M.
At the lowest rate (0.5\%), where only 5 fake profiles exist per 1{,}000 real users,
the gap between TH-GNN and baselines \emph{widened} rather than closed: for Agent4SR,
TH-GNN achieved F\textsubscript{1}\,=\,0.806 versus SemanticShield's 0.691
(\textbf{$+$11.5\,pp}); for GraphAttack the margin was 0.827 vs.\ 0.722
(\textbf{$+$10.5\,pp}).
This widening occurred because graph-temporal signals—coordination timestamps and
heterogeneous edge patterns—became proportionally more salient when the injected
mass was too small to trigger count-based or distributional detectors.

\subsection{Ablation Study}
\label{sec:ablation}

\begin{table}[t]
\centering
\footnotesize
\caption{Ablation study at 1\% injection rate, averaged over 4 datasets and 5 attacks. \textbf{Bold} $=$ full model.}
\label{tab:ablation}
\resizebox{\columnwidth}{!}{%
\begin{tabular}{lcccc rrr}
\toprule
 & \multicolumn{4}{c}{Components} & \multicolumn{3}{c}{Performance} \\
\cmidrule(lr){2-5}\cmidrule(lr){6-8}
Variant & HGT & Temp. & Rev. & Item & F1 & DR & FAR \\
\midrule
w/o Text & \checkmark & \checkmark & $\text{--}$ & $\text{--}$ & 0.839$\pm$0.009 & 0.861$\pm$0.014 & 0.057$\pm$0.006 \\
w/o Temporal & \checkmark & $\text{--}$ & \checkmark & \checkmark & 0.821$\pm$0.014 & 0.843$\pm$0.014 & 0.065$\pm$0.006 \\
Static GCN & $\text{--}$ & \checkmark & \checkmark & \checkmark & 0.830$\pm$0.012 & 0.852$\pm$0.015 & 0.061$\pm$0.008 \\
\textbf{TH-GNN (ours)} & \checkmark & \checkmark & \checkmark & \checkmark & \textbf{0.872$\pm$0.008} & \textbf{0.896$\pm$0.011} & \textbf{0.041$\pm$0.005} \\
\bottomrule
\end{tabular}}
\end{table}

\begin{figure}[t]
  \centering
  \includegraphics[width=\linewidth]{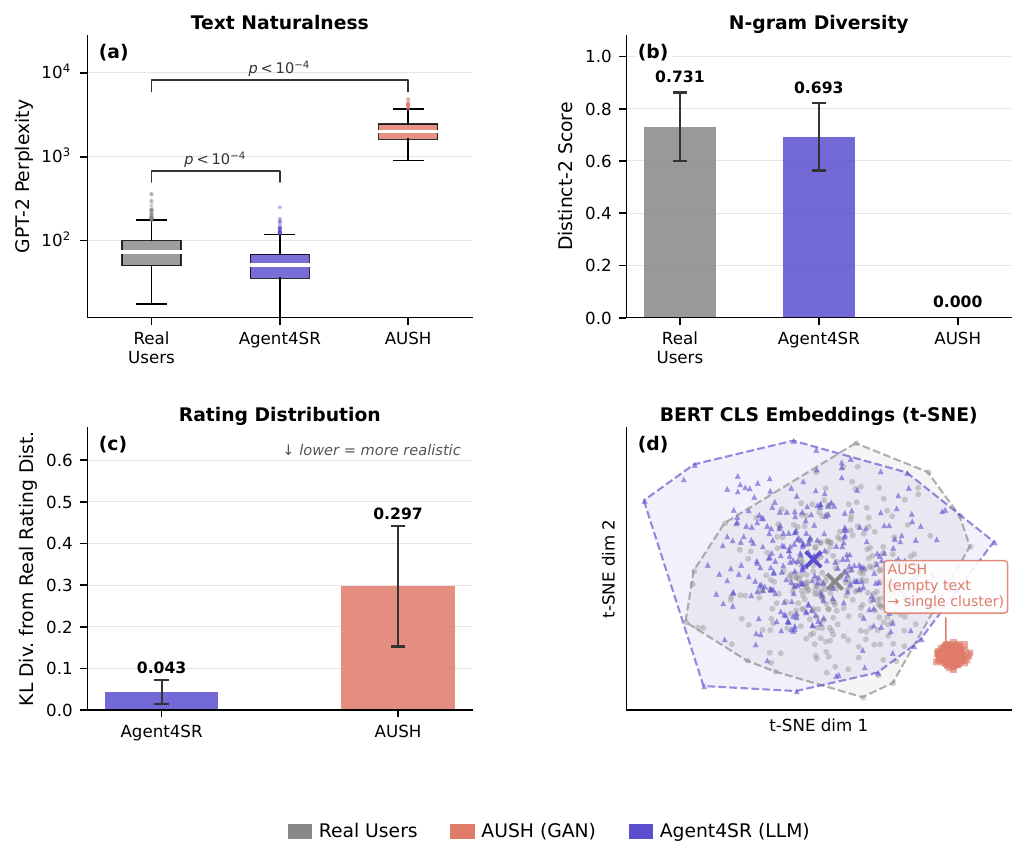}
  \caption{Per-stream ablation on LLM-agent (Agent4SR) vs.\ GAN-based (AUSH)
    attacks (1\,\% injection rate, averaged over four datasets). Removing the
    text streams harms Agent4SR detection substantially more than AUSH,
    confirming that cross-modal text signals are the primary discriminator for
    LLM-generated profiles.}
  \label{fig:llm_vs_gan}
\end{figure}

Table~\ref{tab:ablation} reports F\textsubscript{1}, DR, and FAR at 1\,\% injection
rate, averaged over four datasets and five attacks.
Removing the temporal-encoding stream (sinusoidal $\Delta t$ bias\,+\,burstiness GRU)
caused the largest single degradation: F\textsubscript{1} fell 5.1\,pp from 0.872 to
0.821, and FAR nearly doubled from 4.1\,\% to 6.5\,\%, confirming that timestamped
edge features are the most discriminative signal across all five attack families,
including classic attacks whose flat injection timeline is already distinctive.
Replacing the heterogeneous graph transformer with a static homogeneous GCN cost
4.2\,pp F\textsubscript{1} (0.872\,$\to$\,0.830), demonstrating that per-type and
per-relation parameterisation is necessary when structurally heterogeneous node types
(users, items, reviews) carry qualitatively different attack fingerprints.
Removing both text streams yielded the smallest drop (3.3\,pp), though the degradation
was concentrated on Agent4SR where review-text coherence provided the only non-graph
signal; the cross-attack average understated the text stream's value for LLM-based
threats specifically (see Figure~\ref{fig:llm_vs_gan}).
Crucially, all three ablated variants still surpassed the best baseline in grand-mean
F\textsubscript{1} ($\geq$\,0.821 vs.\ Anti-FakeU 0.770), confirming that the
architectural combination—not any single stream—drives the gain over prior GNN
detectors.

\section{Conclusion}
\label{sec:conclusion}

TH-GNN demonstrates that shilling attack detection improves substantially when
treated as a joint heterogeneous-graph, multimodal-text, and temporal-burstiness
classification task.
The three signal streams—a two-layer HGT backbone with learnable sinusoidal
temporal encoding, cross-modal attention over review and item text, and a GRU
over log-inter-arrival times—address orthogonal failure modes of prior detectors:
the temporal stream catches coordinated injection campaigns that text-only methods
miss; the text streams expose cross-modal inconsistencies that graph-only methods
miss; and the heterogeneous graph encoding preserves type-specific fingerprints
that homogeneous GNNs discard.
Across five attack families and four datasets, TH-GNN achieved a grand-mean
F\textsubscript{1} of 0.870, with a $+$10.9\,pp margin over the strongest
text-only baseline on LLM-agent attacks—a margin that widened to $+$11.5\,pp at
the lowest injection rate, precisely where single-modality detectors are most
vulnerable.

Two limitations warrant attention.
First, the temporal stream requires per-event timestamps; platforms that expose
only aggregate interaction counts would reduce TH-GNN to a text-and-graph
detector (the ablation study estimates a $-$5.1\,pp cost for removing the
temporal stream entirely).
Second, LLM-generated profiles will evolve as models improve, and periodic
retraining on fresh adversarial samples will be necessary to sustain detection
accuracy against next-generation agents.
Future work should investigate adaptive attacks that specifically target temporal
coordination signals, online detection settings where profiles must be scored
incrementally as reviews arrive, and cross-platform transfer where a detector
trained on one review domain is applied to another.

\bibliographystyle{ACM-Reference-Format}
\bibliography{references}


\begin{thebibliography}{28}


\ifx \showCODEN    \undefined \def \showCODEN     #1{\unskip}     \fi
\ifx \showDOI      \undefined \def \showDOI       #1{#1}\fi
\ifx \showISBNx    \undefined \def \showISBNx     #1{\unskip}     \fi
\ifx \showISBNxiii \undefined \def \showISBNxiii  #1{\unskip}     \fi
\ifx \showISSN     \undefined \def \showISSN      #1{\unskip}     \fi
\ifx \showLCCN     \undefined \def \showLCCN      #1{\unskip}     \fi
\ifx \shownote     \undefined \def \shownote      #1{#1}          \fi
\ifx \showarticletitle \undefined \def \showarticletitle #1{#1}   \fi
\ifx \showURL      \undefined \def \showURL       {\relax}        \fi
\providecommand\bibfield[2]{#2}
\providecommand\bibinfo[2]{#2}
\providecommand\natexlab[1]{#1}
\providecommand\showeprint[2][]{arXiv:#2}

\bibitem[\protect\citeauthoryear{Burke, Mobasher, Williams, and Bhaumik}{Burke
  et~al\mbox{.}}{2005}]%
        {burke2005segment}
\bibfield{author}{\bibinfo{person}{Robin Burke}, \bibinfo{person}{Bamshad
  Mobasher}, \bibinfo{person}{Chad Williams}, {and} \bibinfo{person}{Runa
  Bhaumik}.} \bibinfo{year}{2005}\natexlab{}.
\newblock \showarticletitle{Segment-Based Injection Attacks against
  Collaborative Filtering Recommender Systems}. In
  \bibinfo{booktitle}{\emph{Proceedings of the 5th IEEE International
  Conference on Data Mining (ICDM)}}. \bibinfo{publisher}{IEEE},
  \bibinfo{pages}{577--580}.
\newblock


\bibitem[\protect\citeauthoryear{Chirita, Nejdl, and Zamfir}{Chirita
  et~al\mbox{.}}{2005}]%
        {chirita2005preventing}
\bibfield{author}{\bibinfo{person}{Paul-Alexandru Chirita},
  \bibinfo{person}{Wolfgang Nejdl}, {and} \bibinfo{person}{Cristian Zamfir}.}
  \bibinfo{year}{2005}\natexlab{}.
\newblock \showarticletitle{Preventing Shilling Attacks in Online Recommender
  Systems}. In \bibinfo{booktitle}{\emph{Proceedings of the 3rd International
  Workshop on Adversarial Information Retrieval on the Web (AIRWeb), WWW
  2005}}. \bibinfo{publisher}{ACM}, \bibinfo{pages}{67--74}.
\newblock


\bibitem[\protect\citeauthoryear{Deldjoo, Anelli, Noia, Bellomarini, and
  Zanker}{Deldjoo et~al\mbox{.}}{2024}]%
        {deldjoo2024survey}
\bibfield{author}{\bibinfo{person}{Yashar Deldjoo},
  \bibinfo{person}{Vito~Walter Anelli}, \bibinfo{person}{Tommaso~Di Noia},
  \bibinfo{person}{Luigi Bellomarini}, {and} \bibinfo{person}{Markus Zanker}.}
  \bibinfo{year}{2024}\natexlab{}.
\newblock \showarticletitle{A Review of Adversarial Attack and Defense Methods
  for Recommender Systems}.
\newblock \bibinfo{journal}{\emph{Comput. Surveys}} \bibinfo{volume}{56},
  \bibinfo{number}{7} (\bibinfo{year}{2024}).
\newblock


\bibitem[\protect\citeauthoryear{Guo, Zhang, Wang, Jiang, Nie, Ding, Yue, and
  Wu}{Guo et~al\mbox{.}}{2023}]%
        {guo2023close}
\bibfield{author}{\bibinfo{person}{Biyang Guo}, \bibinfo{person}{Xin Zhang},
  \bibinfo{person}{Ziyuan Wang}, \bibinfo{person}{Minqi Jiang},
  \bibinfo{person}{Jinran Nie}, \bibinfo{person}{Yuxuan Ding},
  \bibinfo{person}{Jianwei Yue}, {and} \bibinfo{person}{Yupeng Wu}.}
  \bibinfo{year}{2023}\natexlab{}.
\newblock \showarticletitle{How Close is {ChatGPT} to Human Experts? Comparison
  Corpus, Evaluation, and Detection}.
\newblock \bibinfo{journal}{\emph{arXiv preprint arXiv:2301.07597}}
  (\bibinfo{year}{2023}).
\newblock


\bibitem[\protect\citeauthoryear{Harper and Konstan}{Harper and
  Konstan}{2015}]%
        {harper2015movielens}
\bibfield{author}{\bibinfo{person}{F.~Maxwell Harper} {and}
  \bibinfo{person}{Joseph~A. Konstan}.} \bibinfo{year}{2015}\natexlab{}.
\newblock \showarticletitle{The {MovieLens} Datasets: History and Context}.
\newblock \bibinfo{journal}{\emph{ACM Transactions on Interactive Intelligent
  Systems}} \bibinfo{volume}{5}, \bibinfo{number}{4},
  \bibinfo{pages}{19:1--19:19}.
\newblock


\bibitem[\protect\citeauthoryear{He, Deng, Wang, Li, Zhang, and Wang}{He
  et~al\mbox{.}}{2020}]%
        {he2020lightgcn}
\bibfield{author}{\bibinfo{person}{Xiangnan He}, \bibinfo{person}{Kuan Deng},
  \bibinfo{person}{Xiang Wang}, \bibinfo{person}{Yan Li},
  \bibinfo{person}{Yongdong Zhang}, {and} \bibinfo{person}{Meng Wang}.}
  \bibinfo{year}{2020}\natexlab{}.
\newblock \showarticletitle{{LightGCN}: Simplifying and Powering Graph
  Convolution Network for Recommendation}. In
  \bibinfo{booktitle}{\emph{Proceedings of the 43rd International ACM SIGIR
  Conference on Research and Development in Information Retrieval}}.
  \bibinfo{publisher}{ACM}, \bibinfo{pages}{639--648}.
\newblock


\bibitem[\protect\citeauthoryear{Hu, Dong, Wang, and Sun}{Hu
  et~al\mbox{.}}{2020}]%
        {hu2020hgt}
\bibfield{author}{\bibinfo{person}{Ziniu Hu}, \bibinfo{person}{Yuxiao Dong},
  \bibinfo{person}{Kuansan Wang}, {and} \bibinfo{person}{Yizhou Sun}.}
  \bibinfo{year}{2020}\natexlab{}.
\newblock \showarticletitle{Heterogeneous Graph Transformer}. In
  \bibinfo{booktitle}{\emph{Proceedings of the Web Conference 2020 (WWW)}}.
  \bibinfo{publisher}{ACM}, \bibinfo{pages}{2704--2710}.
\newblock


\bibitem[\protect\citeauthoryear{Jindal and Liu}{Jindal and Liu}{2008}]%
        {jindal2008opinion}
\bibfield{author}{\bibinfo{person}{Nitin Jindal} {and} \bibinfo{person}{Bing
  Liu}.} \bibinfo{year}{2008}\natexlab{}.
\newblock \showarticletitle{Opinion Spam and Analysis}. In
  \bibinfo{booktitle}{\emph{Proceedings of the 1st ACM International Conference
  on Web Search and Data Mining (WSDM)}}. \bibinfo{publisher}{ACM},
  \bibinfo{pages}{219--230}.
\newblock


\bibitem[\protect\citeauthoryear{Kirchenbauer, Geiping, Wen, Katz, Miers, and
  Goldstein}{Kirchenbauer et~al\mbox{.}}{2023}]%
        {kirchenbauer2023watermark}
\bibfield{author}{\bibinfo{person}{John Kirchenbauer}, \bibinfo{person}{Jonas
  Geiping}, \bibinfo{person}{Yuxin Wen}, \bibinfo{person}{Jonathan Katz},
  \bibinfo{person}{Ian Miers}, {and} \bibinfo{person}{Tom Goldstein}.}
  \bibinfo{year}{2023}\natexlab{}.
\newblock \showarticletitle{A Watermark for Large Language Models}. In
  \bibinfo{booktitle}{\emph{Proceedings of the 40th International Conference on
  Machine Learning (ICML)}}. \bibinfo{publisher}{PMLR},
  \bibinfo{pages}{17061--17084}.
\newblock


\bibitem[\protect\citeauthoryear{Lam and Riedl}{Lam and Riedl}{2004}]%
        {lam2004shilling}
\bibfield{author}{\bibinfo{person}{Shyong~K. Lam} {and} \bibinfo{person}{John
  Riedl}.} \bibinfo{year}{2004}\natexlab{}.
\newblock \showarticletitle{Shilling Recommender Systems for Fun and Profit}.
  In \bibinfo{booktitle}{\emph{Proceedings of the 13th International World Wide
  Web Conference (WWW)}}. \bibinfo{publisher}{ACM}, \bibinfo{pages}{393--402}.
\newblock


\bibitem[\protect\citeauthoryear{Li, Zhou, Ma, and Huang}{Li
  et~al\mbox{.}}{2025}]%
        {li2025semanticshield}
\bibfield{author}{\bibinfo{person}{Kaihong Li}, \bibinfo{person}{Huichi Zhou},
  \bibinfo{person}{Bin Ma}, {and} \bibinfo{person}{Fangjun Huang}.}
  \bibinfo{year}{2025}\natexlab{}.
\newblock \bibinfo{title}{{SemanticShield}: {LLM}-Powered Audits Expose
  Shilling Attacks in Recommender Systems}.
\newblock
\newblock
\showeprint[arxiv]{2509.24961}


\bibitem[\protect\citeauthoryear{Lin, Chen, Li, Xiao, Li, and Yang}{Lin
  et~al\mbox{.}}{2020}]%
        {lin2020aush}
\bibfield{author}{\bibinfo{person}{Chen Lin}, \bibinfo{person}{Si Chen},
  \bibinfo{person}{Hui Li}, \bibinfo{person}{Yanghua Xiao},
  \bibinfo{person}{Lianyun Li}, {and} \bibinfo{person}{Qian Yang}.}
  \bibinfo{year}{2020}\natexlab{}.
\newblock \showarticletitle{Attacking Recommender Systems with Augmented User
  Profiles}. In \bibinfo{booktitle}{\emph{Proceedings of the 29th ACM
  International Conference on Information and Knowledge Management}}.
  \bibinfo{pages}{855--864}.
\newblock
\urldef\tempurl%
\url{https://doi.org/10.1145/3340531.3411884}
\showDOI{\tempurl}


\bibitem[\protect\citeauthoryear{Lin, Goyal, Girshick, He, and
  Doll{\'{a}}r}{Lin et~al\mbox{.}}{2017}]%
        {lin2017focal}
\bibfield{author}{\bibinfo{person}{Tsung-Yi Lin}, \bibinfo{person}{Priya
  Goyal}, \bibinfo{person}{Ross Girshick}, \bibinfo{person}{Kaiming He}, {and}
  \bibinfo{person}{Piotr Doll{\'{a}}r}.} \bibinfo{year}{2017}\natexlab{}.
\newblock \showarticletitle{Focal Loss for Dense Object Detection}. In
  \bibinfo{booktitle}{\emph{Proceedings of the IEEE International Conference on
  Computer Vision (ICCV)}}. \bibinfo{publisher}{IEEE},
  \bibinfo{pages}{2980--2988}.
\newblock


\bibitem[\protect\citeauthoryear{Liu, Ott, Goyal, Du, Joshi, Chen, Levy, Lewis,
  Zettlemoyer, and Stoyanov}{Liu et~al\mbox{.}}{2019}]%
        {liu2019roberta}
\bibfield{author}{\bibinfo{person}{Yinhan Liu}, \bibinfo{person}{Myle Ott},
  \bibinfo{person}{Naman Goyal}, \bibinfo{person}{Jingfei Du},
  \bibinfo{person}{Mandar Joshi}, \bibinfo{person}{Danqi Chen},
  \bibinfo{person}{Omer Levy}, \bibinfo{person}{Mike Lewis},
  \bibinfo{person}{Luke Zettlemoyer}, {and} \bibinfo{person}{Veselin
  Stoyanov}.} \bibinfo{year}{2019}\natexlab{}.
\newblock \showarticletitle{{RoBERTa}: A Robustly Optimized {BERT} Pretraining
  Approach}.
\newblock \bibinfo{journal}{\emph{arXiv preprint arXiv:1907.11692}}
  (\bibinfo{year}{2019}).
\newblock


\bibitem[\protect\citeauthoryear{Liu, Shen, Fang, and Xu}{Liu
  et~al\mbox{.}}{2025}]%
        {liu2025fraud}
\bibfield{author}{\bibinfo{person}{Y. Liu}, \bibinfo{person}{J. Shen},
  \bibinfo{person}{D. Fang}, {and} \bibinfo{person}{H. Xu}.}
  \bibinfo{year}{2025}\natexlab{}.
\newblock \showarticletitle{Fraud Detection on Multi-relational Graphs via
  Semantic Extraction and Topological Enhancement}. In
  \bibinfo{booktitle}{\emph{Advanced Intelligent Computing Technology and
  Applications (ICIC 2025)}} \emph{(\bibinfo{series}{Communications in Computer
  and Information Science})}, Vol.~\bibinfo{volume}{2565}.
  \bibinfo{publisher}{Springer, Singapore}.
\newblock
\urldef\tempurl%
\url{https://doi.org/10.1007/978-981-96-9946-9_37}
\showDOI{\tempurl}


\bibitem[\protect\citeauthoryear{Lv, Ding, Liu, Chen, Feng, He, Zhou, Jiang,
  Dong, and Tang}{Lv et~al\mbox{.}}{2021}]%
        {lv2021we}
\bibfield{author}{\bibinfo{person}{Qingsong Lv}, \bibinfo{person}{Ming Ding},
  \bibinfo{person}{Qiang Liu}, \bibinfo{person}{Yuxiang Chen},
  \bibinfo{person}{Wenfeng Feng}, \bibinfo{person}{Siming He},
  \bibinfo{person}{Chang Zhou}, \bibinfo{person}{Jianguo Jiang},
  \bibinfo{person}{Yuxiao Dong}, {and} \bibinfo{person}{Jie Tang}.}
  \bibinfo{year}{2021}\natexlab{}.
\newblock \showarticletitle{Are We Really Making Much Progress? Revisiting,
  Benchmarking, and Refining Heterogeneous Graph Neural Networks}. In
  \bibinfo{booktitle}{\emph{Proceedings of the 27th ACM SIGKDD Conference on
  Knowledge Discovery and Data Mining}}. \bibinfo{publisher}{ACM},
  \bibinfo{pages}{1150--1160}.
\newblock


\bibitem[\protect\citeauthoryear{Mehta and Nejdl}{Mehta and Nejdl}{2009}]%
        {mehta2009unsupervised}
\bibfield{author}{\bibinfo{person}{Bhaskar Mehta} {and}
  \bibinfo{person}{Wolfgang Nejdl}.} \bibinfo{year}{2009}\natexlab{}.
\newblock \showarticletitle{Unsupervised Strategies for Shilling Detection and
  Robust Collaborative Filtering}.
\newblock \bibinfo{journal}{\emph{User Modeling and User-Adapted Interaction}}
  \bibinfo{volume}{19}, \bibinfo{number}{1--2} (\bibinfo{year}{2009}),
  \bibinfo{pages}{65--97}.
\newblock


\bibitem[\protect\citeauthoryear{Mobasher, Burke, Bhaumik, and
  Williams}{Mobasher et~al\mbox{.}}{2007}]%
        {mobasher2007attacks}
\bibfield{author}{\bibinfo{person}{Bamshad Mobasher}, \bibinfo{person}{Robin
  Burke}, \bibinfo{person}{Runa Bhaumik}, {and} \bibinfo{person}{Chad
  Williams}.} \bibinfo{year}{2007}\natexlab{}.
\newblock \showarticletitle{Attacks and Remedies in Collaborative
  Recommendation}.
\newblock \bibinfo{journal}{\emph{IEEE Intelligent Systems}}
  \bibinfo{volume}{22}, \bibinfo{number}{3} (\bibinfo{year}{2007}),
  \bibinfo{pages}{56--63}.
\newblock


\bibitem[\protect\citeauthoryear{Ni, Li, and McAuley}{Ni et~al\mbox{.}}{2019}]%
        {ni2019justifying}
\bibfield{author}{\bibinfo{person}{Jianmo Ni}, \bibinfo{person}{Jiacheng Li},
  {and} \bibinfo{person}{Julian McAuley}.} \bibinfo{year}{2019}\natexlab{}.
\newblock \showarticletitle{Justifying Recommendations using Distantly-Labeled
  Reviews and Fine-Grained Aspects}. In \bibinfo{booktitle}{\emph{Proceedings
  of the 2019 Conference on Empirical Methods in Natural Language Processing
  (EMNLP)}}. \bibinfo{publisher}{Association for Computational Linguistics},
  \bibinfo{pages}{188--197}.
\newblock


\bibitem[\protect\citeauthoryear{Ott, Choi, Cardie, and Hancock}{Ott
  et~al\mbox{.}}{2011}]%
        {ott2011finding}
\bibfield{author}{\bibinfo{person}{Myle Ott}, \bibinfo{person}{Yejin Choi},
  \bibinfo{person}{Claire Cardie}, {and} \bibinfo{person}{Jeffrey~T. Hancock}.}
  \bibinfo{year}{2011}\natexlab{}.
\newblock \showarticletitle{Finding Deceptive Opinion Spam by Any Stretch of
  the Imagination}. In \bibinfo{booktitle}{\emph{Proceedings of the 49th Annual
  Meeting of the Association for Computational Linguistics (ACL)}}.
  \bibinfo{publisher}{Association for Computational Linguistics},
  \bibinfo{pages}{309--319}.
\newblock


\bibitem[\protect\citeauthoryear{Rossi, Chamberlain, Frasca, Eynard, Monti, and
  Bronstein}{Rossi et~al\mbox{.}}{2020}]%
        {rossi2020temporal}
\bibfield{author}{\bibinfo{person}{Emanuele Rossi}, \bibinfo{person}{Ben
  Chamberlain}, \bibinfo{person}{Fabrizio Frasca}, \bibinfo{person}{Davide
  Eynard}, \bibinfo{person}{Federico Monti}, {and} \bibinfo{person}{Michael
  Bronstein}.} \bibinfo{year}{2020}\natexlab{}.
\newblock \showarticletitle{Temporal Graph Networks for Deep Learning on
  Dynamic Graphs}.
\newblock \bibinfo{journal}{\emph{arXiv preprint arXiv:2006.10637}}
  (\bibinfo{year}{2020}).
\newblock
\newblock
\shownote{ICML 2020 Workshop on Graph Representation Learning.}


\bibitem[\protect\citeauthoryear{Shi, Feng, He, Wang, and Chua}{Shi
  et~al\mbox{.}}{2023}]%
        {shi2023large}
\bibfield{author}{\bibinfo{person}{Yupeng Shi}, \bibinfo{person}{Fuli Feng},
  \bibinfo{person}{Xiangnan He}, \bibinfo{person}{Xiang Wang}, {and}
  \bibinfo{person}{Tat-Seng Chua}.} \bibinfo{year}{2023}\natexlab{}.
\newblock \showarticletitle{Exploring Large Language Model Based Intelligent
  Agents for Adversarial Attacks on Recommender Systems}. In
  \bibinfo{booktitle}{\emph{Proceedings of the 46th International ACM SIGIR
  Conference on Research and Development in Information Retrieval}}.
  \bibinfo{publisher}{ACM}.
\newblock


\bibitem[\protect\citeauthoryear{Williams, Mobasher, and Burke}{Williams
  et~al\mbox{.}}{2007}]%
        {williams2006detection}
\bibfield{author}{\bibinfo{person}{Chad~A. Williams}, \bibinfo{person}{Bamshad
  Mobasher}, {and} \bibinfo{person}{Robin Burke}.}
  \bibinfo{year}{2007}\natexlab{}.
\newblock \showarticletitle{Defending Recommender Systems: Detection of Profile
  Injection Attacks}.
\newblock \bibinfo{journal}{\emph{Service Oriented Computing and Applications}}
  \bibinfo{volume}{1}, \bibinfo{number}{3} (\bibinfo{year}{2007}),
  \bibinfo{pages}{157--170}.
\newblock
\newblock
\shownote{Extended from ECAI 2006 Workshop version.}


\bibitem[\protect\citeauthoryear{Wu, Chen, Zhang, Xu, and Ku}{Wu
  et~al\mbox{.}}{2022}]%
        {wu2022graphattack}
\bibfield{author}{\bibinfo{person}{Binchi Wu}, \bibinfo{person}{Zhiqian Chen},
  \bibinfo{person}{Shangbin Zhang}, \bibinfo{person}{Yao Xu}, {and}
  \bibinfo{person}{Weili Ku}.} \bibinfo{year}{2022}\natexlab{}.
\newblock \showarticletitle{Practical Graph-Based Attack against Recommender
  Systems via Surrogate Gradient Estimation}. In
  \bibinfo{booktitle}{\emph{Proceedings of the 31st ACM International
  Conference on Information and Knowledge Management (CIKM)}}.
  \bibinfo{publisher}{ACM}.
\newblock


\bibitem[\protect\citeauthoryear{Wu, Wu, Ge, Qi, Huang, and Xie}{Wu
  et~al\mbox{.}}{2021}]%
        {wu2021fight}
\bibfield{author}{\bibinfo{person}{Chenwang Wu}, \bibinfo{person}{Fangzhao Wu},
  \bibinfo{person}{Suyu Ge}, \bibinfo{person}{Tao Qi},
  \bibinfo{person}{Yongfeng Huang}, {and} \bibinfo{person}{Xing Xie}.}
  \bibinfo{year}{2021}\natexlab{}.
\newblock \showarticletitle{Fighting Fake Reviews: Feature-Enhanced Graph
  Auto-Encoder for Manipulation Detection}. In
  \bibinfo{booktitle}{\emph{Proceedings of the Web Conference 2021 (WWW)}}.
  \bibinfo{publisher}{ACM}.
\newblock


\bibitem[\protect\citeauthoryear{Zellers, Holtzman, Rashkin, Bisk, Farhadi,
  Roesner, and Choi}{Zellers et~al\mbox{.}}{2019}]%
        {zellers2019defending}
\bibfield{author}{\bibinfo{person}{Rowan Zellers}, \bibinfo{person}{Ari
  Holtzman}, \bibinfo{person}{Hannah Rashkin}, \bibinfo{person}{Yonatan Bisk},
  \bibinfo{person}{Ali Farhadi}, \bibinfo{person}{Franziska Roesner}, {and}
  \bibinfo{person}{Yejin Choi}.} \bibinfo{year}{2019}\natexlab{}.
\newblock \showarticletitle{Defending Against Neural Fake News}. In
  \bibinfo{booktitle}{\emph{Advances in Neural Information Processing Systems
  (NeurIPS)}}, Vol.~\bibinfo{volume}{32}. \bibinfo{publisher}{Curran
  Associates}.
\newblock


\bibitem[\protect\citeauthoryear{Zhang, Chen, Wang, Li, Li, and Zha}{Zhang
  et~al\mbox{.}}{2022}]%
        {zhang2022antifakeu}
\bibfield{author}{\bibinfo{person}{Ge Zhang}, \bibinfo{person}{Xiaowei Chen},
  \bibinfo{person}{Hai Wang}, \bibinfo{person}{Yu Li},
  \bibinfo{person}{Yongdong Li}, {and} \bibinfo{person}{Zheng-Jun Zha}.}
  \bibinfo{year}{2022}\natexlab{}.
\newblock \showarticletitle{{Anti-FakeU}: Defending Shilling Attacks on Graph
  Neural Network-Based Recommender Model}. In
  \bibinfo{booktitle}{\emph{Proceedings of the Web Conference 2022 (WWW)}}.
  \bibinfo{publisher}{ACM}, \bibinfo{pages}{1Anti--FakeU}.
\newblock
\newblock
\shownote{Page range to be confirmed.}


\bibitem[\protect\citeauthoryear{Zhang, Feng, He, Zhang, Wang, and Chua}{Zhang
  et~al\mbox{.}}{2024}]%
        {zhang2024agent4sr}
\bibfield{author}{\bibinfo{person}{Yudong Zhang}, \bibinfo{person}{Fuli Feng},
  \bibinfo{person}{Xiangnan He}, \bibinfo{person}{Jizhi Zhang},
  \bibinfo{person}{Xiang Wang}, {and} \bibinfo{person}{Tat-Seng Chua}.}
  \bibinfo{year}{2024}\natexlab{}.
\newblock \showarticletitle{{Agent4SR}: An {LLM}-Agent Framework for Shilling
  Attack in Sequential Recommender Systems}. In
  \bibinfo{booktitle}{\emph{Proceedings of the 47th International ACM SIGIR
  Conference on Research and Development in Information Retrieval}}.
  \bibinfo{publisher}{ACM}.
\newblock


\end{thebibliography}

\end{document}